\documentclass[runningheads]{llncs}

\usepackage{graphicx}
\usepackage{xcolor}
\usepackage{amsmath}
\usepackage{subcaption}
\usepackage{multirow}
\definecolor{darkcyan}{rgb}{0.0, 0.55, 0.55}
\definecolor{navy}{rgb}{0.0, 0, 0.5}
\usepackage[
        destlabel=true,
        colorlinks=true,
        urlcolor=darkcyan,
        linkcolor=navy,
        citecolor=black]{hyperref} 
\usepackage{arydshln}
\usepackage{mathtools}

\begin{document}

\title{
Energy-Efficient Control of Cable Robots Exploiting Natural Dynamics and Task Knowledge 
}
\titlerunning{
Energy-Efficient Control of Cable Robots 
}

\author{Boris Deroo\inst{1}
\orcidID{0000-0002-9089-9341}        
        \and Erwin Aertbeli\"en \inst{1,2}
\orcidID{0000-0002-4514-0934}
        \and Wilm Decr\'e \inst{1,2}
\orcidID{0000-0002-9724-8103}
        \and Herman Bruyninckx \inst{1,2,3}
\orcidID{0000-0003-3776-1025}
        }

\institute{KU Leuven, Department of Mechanical Engineering, Belgium \\
\email{first\_name.second\_name@kuleuven.be}\\
\and
Flanders Make, Leuven, Belgium 
\and
TU Eindhoven, Mechanical Engineering, The Netherlands
}

\authorrunning{B. Deroo et al.} 

\maketitle 

\newcommand*\circled[1]{
        \!\!\!\raisebox{.5pt}{\textcircled{\raisebox{-.5pt}{\footnotesize{#1}}}}}\!
\newcommand*\circledtable[1]{
        \raisebox{-.3pt}{\textcircled{\raisebox{-.65pt}{\scriptsize{#1}}}}} 
\newcommand*\figurewidth{0.7\linewidth}

\vspace{-8mm}

\begin{abstract}
This paper focusses on the energy-efficient control of a cable-driven
robot for 
tasks that only require precise positioning at few points in their
motion, and where that accuracy can be obtained through contacts. This
includes the majority of pick-and-place operations.

Knowledge about the task is directly taken into account when specifying 
the control execution.
The natural dynamics of the system can be exploited when there is a 
tolerance on the position of the trajectory. 
Brakes are actively used to replace standstill torques, and as passive 
actuation.
This is executed with a hybrid discrete-continuous controller. 
A discrete controller is used to specify and coordinate between subtasks,  
and based on the requirements of these specific subtasks, specific, robust, 
continuous controllers are constructed. This approach allows for less stiff 
and thus saver, and cheaper hardware to be used.
For a planar pick-and-place operation, it was found that this results 
in energy savings of more than $30\%$. However, when the payload 
moves with the natural dynamics, there is less control of the followed 
trajectory and its timing compared to a traditional trajectory-based 
execution.
Also, the presented approach implies a fundamentally different way to specify 
and execute tasks.

\end{abstract}

\section{Introduction}    \label{sec:introduction}
Traditionally the execution of robot manipulator tasks is not 
focussed on limiting the energy consumption, but rather on speed and 
precision. 
This often results in robots that are more precise, and thus 
stiffer, heavier, and more
expensive, than strictly necessary for the task.
In addition, increasing energy costs and a growing need for more 
sustainability push for more energy-efficient solutions.
Practical applications of tasks that can also be accomplished with 
less accurate robots are numerous in industry, e.g. palletising, truck 
unloading, box stacking, etc. 
These types of pick-and-place tasks only require high precision 
at the start and end of the execution, but not for the gross of
the motion.
These tasks are the main focus of this paper and will 
henceforth be referred to as `transportation tasks'.

\begin{figure}[ht]
    \vspace{-4mm}
    \centering
    \includegraphics[width = 6.4cm]{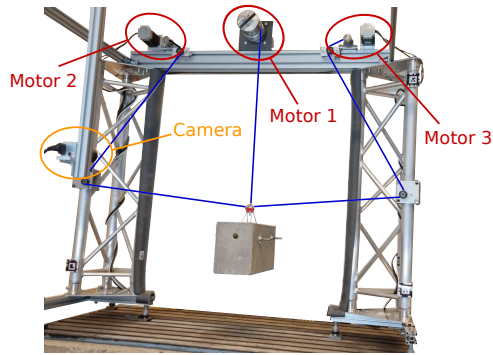}
    \caption{Cable-driven parallel robot that was used in this work. 
            The actuated cables are highlighted with blue lines.
            All motors are equipped with a brake.}
    \label{fig:setup}
    \vspace{-4mm}
\end{figure}

This paper demonstrates that by adapting the definition of a task 
in a smart way, by utilising knowledge about the system and task directly
in the control execution,  
a more energy-efficient control can be achieved using simple control 
strategies that do not require accurate modelling nor a high 
computational load. 
To illustrate these strategies, a robot manipulator was built to 
manipulate relatively high payloads of 10 to 100kg (Fig. \ref{fig:setup}). 
For the majority of the task execution, high precision 
is not necessary, and natural constraints \cite{Eppner2015} can be 
used to achieve the required precision. 
By using cables as actuation, the moving mass of the 
robot is minimal, and lowering the 
inherent energy consumption. 

\noindent
Research covering non-stiff manipulators for such applications
typically attempts to mimic the general purpose robot requirements 
and applications, actively trying to eliminate vibrations induced by 
its natural dynamics \cite{Malzahn2014}. This work argues that a 
better way is to compete on specific application use cases by 
making them more energy efficient, and 
not attempt to replace industrial robots altogether.

The energy consumption can be minimised  
by optimising the executed trajectory, by optimising the 
execution time \cite{Pellicciari2013}, or the followed path in 
space \cite{Paes2014}. However, the robot is still constrained 
to follow a point-to-point (PTP) trajectory that might 
differ from the natural dynamics. 

Attaching springs to a SCARA robot in order to store energy in between 
cycles of a task has shown to significantly increase
the energy efficiency \cite{Goya2012}.
The natural dynamics of a robot with series elastic actuators (SEA) were 
also used to achieve end effector velocities that were much higher than 
a stiff robot \cite{Haddadin2012}, limiting the required power of the motors.
While these works show promising results, the concept of moving with the
natural dynamics has not been explored extensively in robotic manipulation.

To hold a constant position during control, usually a standstill torque is 
generated in the actuators, continuously consuming power.    
Instead brakes could be used to achieve the same effect.
Commercial robots are already equipped with normally-on brakes, 
consuming a constant power to deactivate the brakes. 
Thus, utilising the brakes in the control has a double positive 
effect on the energy consumption. 
The downside being that it introduces a time delay to (de)activate 
them, and the user usually cannot individually control the brakes 
of an industrial robot.
Brakes have already been used to passively control robots
that experience an external force
\cite{Hirata2011,Andreetto2016}. 
Braking a joint implies a geometric constraint, discretely changing
the natural dynamics of the system. 
This constraint can be easily incorporated in the kinematic control. 

A common challenge in cable robots is the redundancy resolution 
\cite{Lamaury2013,Oh2007}. As will be explained in section 
\ref{sec:continuous_control}, the method described 
in this work will solve this by selectively disabling the motors.

The task context associated with the application determines 
the requirements and constraints throughout the task execution.
For transportation tasks, these requirements and 
constraints are subject to change during the operation. 
For example, the placement or insertion of a payload typically 
requires a higher precision than the transportation in free space.  
This knowledge can be taken into account to split the task into 
multiple simple subtasks with different requirements.
For these subtasks, specific controllers and mechatronics can be 
developed that focus on their robust execution, and 
their specific requirements.
Monitors can be used to track the trends of the continuous 
execution, and coordinate the discrete switching between the controllers.
Thus, for flexible robots
it makes sense to focus on such tasks 
that do not have strict precision requirements, or where this 
precision does not need to originate directly from the control.

Summarising, the contributions of this paper are the following: 
\begin{itemize}
\item 
        Exploiting the natural dynamics of the system for higher 
        energy-efficiency task execution, while still fulfilling the 
        required position precision of the task. 
        
\item 
        Using insights about the task and the mechanical 
        system to develop simple, robust continuous controllers  
        with realtime task execution monitors that feed into  a discrete task execution.

\item
        Achieving the required task precision by making use of natural 
        constraints of the environment, or artificially induced constraints 
        on the cable lengths by means of braking. 
\end{itemize}

\section{Task Specification}    \label{sec:task_specification}

As specified earlier, the methodology focusses on a transportation task.
The goal is to place a payload next to a previously placed payload, or 
mechanical constraint, as illustrated in Fig. 
\ref{fig:discrete_states}.

\subsection{Assumptions and Knowledge}

The task specification of the use case investigated in this work
makes following assumptions:
\begin{itemize}
    \item The payload is not fragile and the environment can be used to 
    mechanically dampen vibrations, without damaging the payload or the 
    environment.
    
    \item The world model information, such as the position of 
    objects already placed in the workspace, is known to the controller.
    
    \item Motor brakes can be individually activated and controlled.
    
    \item The top motor is positioned over the previously placed 
    payload, such that the current payload can be swung over it.

    \item Grasping the payload is out of the scope of this paper
\end{itemize}  

\noindent The control approach makes use of the following 
knowledge in the task specification: 
\begin{itemize}
    \item The payload has a significant mass, thus
    gravitational force can be used to ensure cable tension. 

    \item The execution can be split up in a lifting, transporting, 
    dropping and (optional) fine-positioning state.
\end{itemize}

\subsection{Energy measures}
The total electric power consumed by a single motor is given by: 
\begin{equation}
\begin{gathered}
    P_{el} = P_{mech} + P_{mech,loss} + P_{el,loss} 
    = V I_a,
\end{gathered}
\end{equation}  
with $P_{mech}$ the mechanical power, 
$P_{mech,loss}$ the mechanical losses due to friction in the motor and 
transmission, $P_{el,loss}$ the electrical losses of the motor, $V$ the 
voltage applied to the motor, and $I_a$ the armature current. 

The electrical losses are typically dominated by the copper losses 
$P_{cu,loss}$, determined by the motor resistance $R_a$ and armature
current $I_a$: 
\begin{equation}
    \begin{gathered}
        P_{cu,loss} = R_a^{} I_{a}^{2}.
    \end{gathered}
\end{equation}
The energy consumption can be found by integration of the
power over time $t$:
\begin{equation}
\label{eq:el_energy}
    E_{el} = \int_{0}^{t} P_{el} \mathrm{d}t.
\end{equation}

\subsection{Approach}
This paper describes three concepts with the aim 
of a robust, energy-efficient robot control. 
While these concepts are rather straightforward and intuitive, they 
are not yet utilised in mainstream manipulator control.
It is not argued that all of these principles 
need to be used, but if the (sub)task allows it, using any 
of these principles can have a positive impact on the energy-efficiency.

\vspace{-4mm}
\subsubsection{Contextual prior knowledge:}
When humans perform a transportation task, they do  
not perform strict PTP motions such as traditional industrial 
robots. Instead, movements with a certain tolerance on the position 
are performed. 
This allows the natural dynamics of the system to be exploited, 
as will be explained in the following subsection. Typically, the 
tolerances come from 
knowledge about both the environment and the task context. For example, 
the spatial constraints, 
fragility of the payload, 
if a certain part of the task requires a higher precision, 
etc. 
It is clear that this knowledge precedes the task execution and 
determines how the human will perform the task.
The execution is generally done in multiple states, e.g., picking up 
the payload, moving and placing near the target position, making small 
adjustments when necessary.

This knowledge is used to split up the task in multiple 
subtasks and identify the different requirements. Robust 
controllers and monitors are then developed to perform and coordinate 
between these subtasks. 
Examples of such requirements are crane like operations such as:
 lifting the load to a certain 
height, transporting it without colliding, and lowering the load 
until contact is made.

The task also does not require high control precision throughout, 
but only for the initial grasping and final placement. 
In addition, this does not need to come only from the 
control.
Geometric constraints such as the environment or a previously placed
payload can be used to achieve this accuracy by sliding against them. 
This is further explained in section \ref{sec:discrete_control}.

By using this knowledge, lower-cost (and often also lower-weight) 
hardware can be used, so that a more robust, 
energy efficient execution can be developed. 
Thus, for a repetitive task, the cost of 
designing and implementing a task-specific controller is not 
necessarily higher than a generic, less energy-efficient controller.\\

\vspace{-8mm}
\subsubsection{Exploiting natural dynamics}
In this work, the natural dynamics of the system are used to inject 
as little energy as possible, resulting in energy-efficient 
motions.
However, precise control of the timing is lost when the system freely
follows its natural dynamics.

Due to the layout of the used cable robot (Fig.1), when the end effector is
in a fully constrained position, releasing the power of one (or more) 
of the motors, will result in a pendulum-like swing around the 
cables that are still powered, or braked. 
This swing is used in the control strategy to cover the horizontal 
distance while consuming a minimal amount of energy. 

\vspace{-4mm}
\subsubsection{Active use of brakes}
Based on the context, certain subtasks may occur where a joint 
does not need to move. Instead of producing a constant standstill
torque, it can also be opted to brake the joint.
Another case occurs when the demanded motion is in line with 
external forces such as gravity. In case of a continuous brake,
the brake force can be directly controlled to achieve a certain 
resulting force. 
With a discrete brake, a tolerance region can be determined between  
which the brake switches on-and-off to achieve a similar effect.
Section \ref{sec:continuous_control} utilises this concept to drop the
payload without driving the motor. The brakes can also be used to stop 
the natural dynamics, if necessary.
\section{Control Strategy} \label{sec:control_strategy}
The controller is of the hybrid continuous-discrete type.
The task is split up in subtasks which are executed 
with specific continuous controllers. Monitors are used to 
trigger transitions in the discrete control, implemented 
as a finite state machine (FSM). 

\subsection{Continuous control} \label{sec:continuous_control}
\begin{figure}[h]
    \vspace{-4mm}
    \centering
    \includegraphics[width = 5.5 cm]{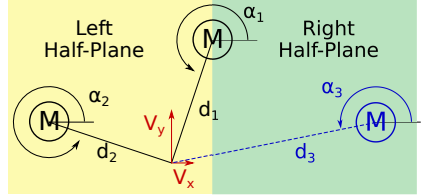}
    \caption{Kinematics of a planar, redundant cable robot. In the 
    depicted scenario,  
    the right actuator (depicted in blue) is not being driven.}
    \label{fig:InverseKinematics}
    \vspace{-4mm}
\end{figure}

At the lowest level, each of the motors is 
either controlled by a velocity PID, or a current controller. 
The latter offering more opportunities for energy savings, at the cost 
of a higher control design effort.
Depending on the subtask and the joint, inverse kinematics are 
used to construct a Cartesian velocity controller, 
or a joint current controller.
The layout of the cable robot is illustrated in Fig. 
\ref{fig:InverseKinematics}. 

\vspace{-6mm}
\subsubsection{Cartesian controller}
The top motor act as a hoist, and is situated such that it can 
always deliver the majority of the gravity compensating force,
and hence can take the role of a hoist.
Depending on which side of the vertical plane (Fig. 
\ref{fig:InverseKinematics}) the payload is situated, only the side motor that is 
in the same half plane will be able to deliver a force that can 
counteract gravity. The other one is not used for the 
manipulation.
This reduces the redundancy to zero, meaning that a unique 
inverse kinematic solution exists, with two complementary discrete 
modes. 

The control input of the robot is the velocity of the non-redundant 
motors which 
is related to the rate of change of their cable length 
$\dot{d_i} = n_i r_i \dot{\theta_i}$. With $n_i$, $r_i$, and
$\dot{\theta_i}$ respectively the gear ratio, the drum radius, and 
the velocity of motor $i$.
In the following, the symbol $x_j$ and $x_j'$ signify parameter $x$ 
of respectively the driven and non-driven side motor.
The Jacobian can easily be derived from the kinematics. The resulting 
inverse kinematic equations are given by:

\begin{equation}
    \begin{bmatrix}
        \dot{d_1} \\ 
        \dot{d_j} 
    \end{bmatrix}
    = 
    \begin{bmatrix}
        \dfrac{-\mathrm{sin}(\alpha_j)}{\mathrm{sin}(\alpha_1-\alpha_j)} 
        & \dfrac{\mathrm{sin}(\alpha_1)}{\mathrm{sin}(\alpha_1-\alpha_j)} \\[14pt]
        \dfrac{\mathrm{cos}(\alpha_j)}{\mathrm{sin}(\alpha_1-\alpha_j)}
        & \dfrac{-\mathrm{cos}(\alpha_1)}{\mathrm{sin}(\alpha_1-\alpha_j)}
    \end{bmatrix}^{-1}
    \begin{bmatrix}
        V_x \\
        V_y
    \end{bmatrix}
    ,
\end{equation}
%
when the movement is in the `push' direction of the driven side motor 
($\dot{l_i}>0$), gravity is used as the driving force of the 
motion. 
To ensure that the non-driven side cable does not slack, a  
current, just slightly larger than the static friction, is 
maintained in the non-driven side motor when the motion is along 
the $-\dot{d_j}'$ direction. Otherwise gravity ensures cable tension, and 
the motor is not powered.

\vspace{-2mm}
\subsubsection{Current controller}
The current controller is used when one of the actuators is braked.
Braking an actuator implies a geometrical
constraint, such that the end effector has to be on a spherical 
surface with a radius that is determined by the cable length. 
This reduces the mobility of the end effector to 1 DoF in a plane, 
where the other joints can be used to move along the circular constraint. 
This results in simple kinematics (assuming the braked cable 
remains tensioned).

\begin{figure}[h]
    \vspace{-4mm}
    \centering
    \includegraphics[width = 5.2cm]{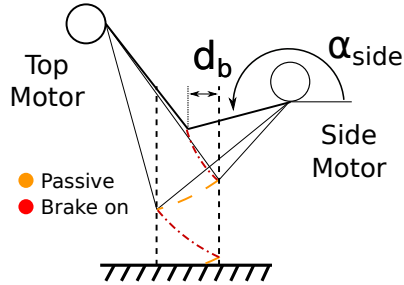}
    \caption{Dropping the payload by passive control. Red circle segments
    indicate the side motor is braked. Orange segments indicate that the 
    side motor is passive, allowing the cable to move freely due to gravity.}
    \label{fig:BrakeDrop}
    \vspace{-4mm}
\end{figure}

This method is used when the desired motion is in the direction of 
gravity, and thus the control can occur passively. E.g., when dropping 
the payload the side motor brake is controlled to avoid a holding 
torque (Fig. \ref{fig:BrakeDrop}). 

\vspace{-6mm}
\subsection{Discrete control}   \label{sec:discrete_control}
\begin{figure}[h]
    \vspace{-2mm}
    \centering
    \begin{subfigure}[b]{0.542\linewidth}
        \centering
        \includegraphics[height=3.5cm]{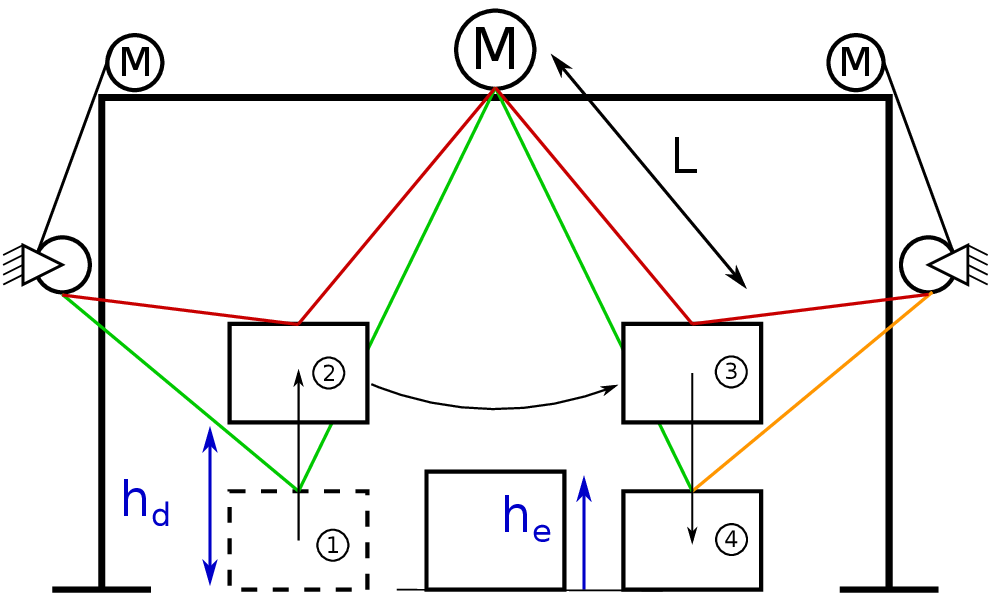}
    \end{subfigure}
    \begin{subfigure}[b]{0.45\linewidth}
        \centering
        \includegraphics[height=3.5cm]{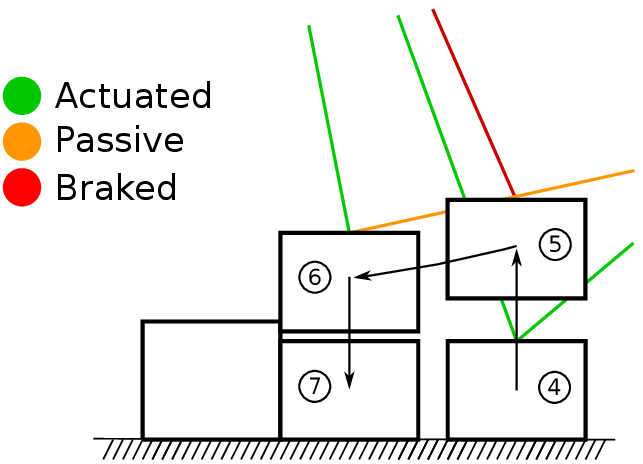}
    \end{subfigure} 
    \caption{Different subtasks to be executed. The arrows indicate the motion 
        in each subtask. The numbers 
        indicate when a monitor triggers that indicates the end 
        of the current subtask. 
        Cables depicted in green are actuated, in orange can passively 
        move, in red are constrained by a brake.  
        }
    \vspace{-4mm}
    \label{fig:discrete_states}
\end{figure}

The aforementioned continuous control is implemented for each of the 
specific subtasks by means of a higher level discrete controller. 
Figure \ref{fig:discrete_states} illustrates the different subtasks. 
Each circled number represents when the end criteria monitor of one of 
the discrete states should trigger, which serves as a signal for the 
discrete controller to go to the following state.      
The high level control consists of the states illustrated in Fig. 
\ref{fig:discrete_states}

\vspace{-8mm}
\subsubsection{Lifting state:}
Monitor \circled{1} indicates the system is operational and the first subtask can be executed.
The load is lifted upwards with a 
feed-forward Cartesian velocity input.   
When the end effector has reached a certain height $h_d$, monitor 
\circled{2} 
triggers, signalling the discrete controller to go to the `swing state'.
$h_d$ is such that the payload does not collide with previously 
placed payloads, nor the ground during the swing state. 
This height can be determined by, e.g., a camera system and 
is assumed to be known in this paper. 
During this state, energy is injected into the payload and stored as 
gravitational energy. 

\vspace{-2mm}
\subsubsection{Swing state}
Releasing the side motor power causes the payload to execute a free 
fall. However, the top motor brake is  activated, 
constraining the motion along a circle, resulting in a pendulum-like motion. 
During this motion, the direction of the end position is monitored. 
After half a period, when the end effector reaches its apex, 
monitor \circled{3} triggers. This causes the brake of the other side 
motor to activate, holding the payload steady. 
A small amount of energy is consumed by a single 
side motor to maintain cable tension. By keeping the brake of the  
top motor active, no holding torque (and thus no energy) is required.

\vspace{-2mm}
\subsubsection{Drop state}
After the swing, the load is dropped to the ground in the vicinity 
of the target position by applying braking actions (Fig. \ref{fig:BrakeDrop}).
During the dropping motion, the current of the top motor is monitored.
A sudden discrepancy  
of this value can be used to detect an impact force. This is used to 
detect the collision when the payload touches the ground, triggering 
monitor \circled{4}. 
Which in turn switches to the fine-positioning state.

\vspace{-2mm}
\subsubsection{Fine-positioning state}
A sequence of actions is performed to place the payload at the target 
position. First the load is slightly lifted upwards until the cable 
length $l_2$ is such that the payload can not collide with the ground. 
This triggers monitor
\circled{5}. 
Which causes the top motor to brake and the side 
motor to release power, causing a free swing motion, at very low speed 
until the payload collides with the 
previously placed payload, triggering monitor
\circled{6}.
Afterwards the payload is moved straight down with a Cartesian controller, 
until it collides with the ground triggering monitor
\circled{7}, terminating the fine-positioning and the task execution.
To lift the payload a small amount of energy is injected, which is also true 
for the final drop since this is done with active actuation. 
 
\section{Experimental Set-up}          \label{sec:experiment_setup}
These concepts were validated on 
a parallel cable-driven robot (Fig. \ref{fig:setup}). 
The system is able to manipulate a payload of $14$kg
in the vertical plane, with a workspace of $1\times 1$m 
through a 3 DoF redundant actuation. 

A $750$W Beckhoff AM8032 motor with a gear ratio of 70, controlled
by a Beckhoff AX5203 drive, is positioned at the top such that it 
mostly compensates the gravitational force of the payload. 
It has a normally-on electromagnetic brake, allowing discrete braking 
actions. A power of $P_{b,top}=11$W is required to deactivate the brake.
The cable is clamped and wound around a $4.2$cm radius drum that is 
connected at the shaft output. 

\begin{figure}[h]
    \vspace{-4mm} 
    \centering
        \includegraphics[width = 0.58\linewidth]{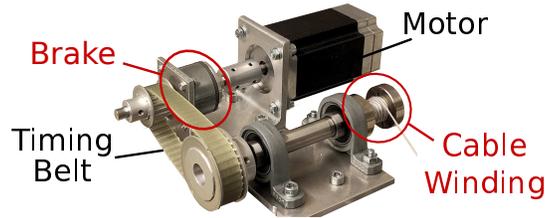}
    \caption{BLDC motor, brake and timing belt transmission 
        used for the side actuation.}
    \vspace{-4mm}
    \label{fig:bldc}
\end{figure}

The payload is connected at both sides to a $188$W BLDC motor 
(Fig. \ref{fig:bldc}). Both motors are controlled through a VESC 
drive, an open-source hardware project \cite{VESC}. 
The motors have an internal gear ratio of $8$, and are 
connected by a timing belt and pulley system with an additional 
gear ratio of $30/16$ to a rotary shaft with a radius of $8$mm.
The cables are directly wound around this shaft, and guided along 
pulleys towards the side of the payload (Fig. \ref{fig:setup} and 
\ref{fig:bldc}).
The motors are equipped with normally-off electromagnetic 
brakes, instead of normally-on due to long supplier lead times on 
the latter. However, in an industrial application the robot
would be equipped with normally-on brakes. Thus, the brakes will 
be treated as normally-on during the energy consumption analysis.
A power of $P_{b,left}=P_{b,right}=8$W
is necessary to activate the brakes.

\noindent
All the used cables are made from Dyneema SK78 with a diameter of $1$mm. 
These cables are lightweight, have a low working stretch ($<1\%$), 
and can carry up to $1.95$kN. 
The current $I_a$ and voltage $V$ supplied to the motors are 
directly obtained from the drives. 
The control feedback relies on the position of the end effector. 
In most cable robots this can be derived from the encoder 
positions, but since cable tension cannot be ensured during the 
free swing of the payload, a camera with a sensor size of 
$1920\times 1280$p is used. 
This allows direct end position tracking of the end effector at 
a rate of $150$Hz with a resolution of $\pm 1$mm.  
\section{Experimental Results}                   \label{sec:results}
The proposed approach was experimentally compared against 
a sequence of PTP motions, each with a trapezoidal velocity profile.

\begin{figure}[h!]
    \centering
    \includegraphics[width = 0.68\linewidth]{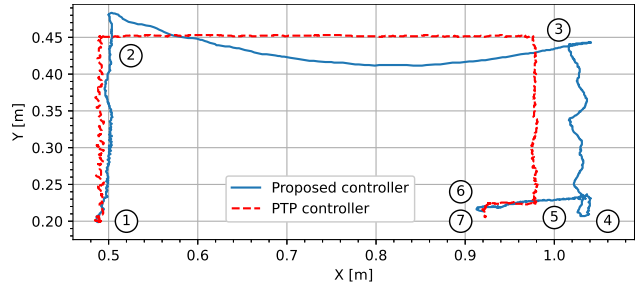}
    \caption{Executed trajectory of the proposed and the PTP controller.}
    \label{fig:trajectories}
    \vspace{-6mm}
\end{figure}

The trajectories corresponding to both methods are shown in 
Fig. \ref{fig:trajectories}.
The PTP-trajectory performs a vertical lift up, followed by a horizontal 
transportation, a vertical drop slightly above the ground, and finally a 
horizontal and vertical movement that finishes the placement. 
Parts of the trajectory that are similar 
to the end criteria of the proposed method are marked with 
circled numbers in the figure
The PTP-trajectory does not make use of a contact with the ground to 
detect when the drop has ended, it goes straight from 
\circled{3} to \circled{5}.

It should be noted that the side motors are not intended for servo 
applications, causing jittery motions at low speeds.
This caused a chattering effect when the PTP controller was used, 
since an exact position needs to be followed, resulting in higher 
jerks (e.g. between \circled{1} and \circled{2}).
The proposed controller does not rely on position tracking, 
and as such does not induce this chattering.

\begin{figure*}[h!]
    \vspace{-4mm}
    \centering
    \begin{subfigure}[b]{0.455\linewidth}
        \centering
        \includegraphics[width = \textwidth]{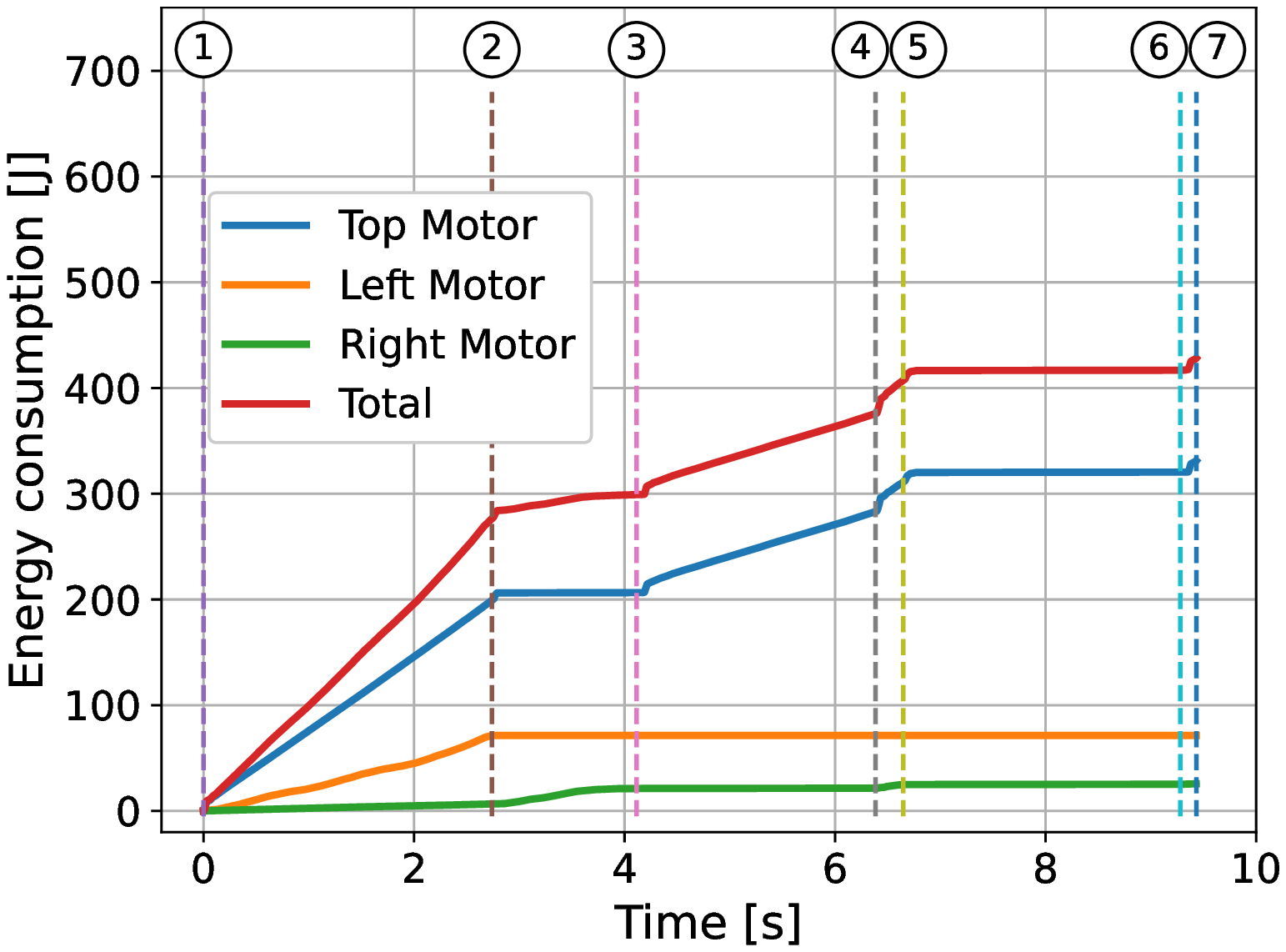}
        \caption{Proposed controller}
        \label{fig:energy_swing}
    \end{subfigure} 
    \begin{subfigure}[b]{0.455\linewidth}
        \centering
        \includegraphics[width = \textwidth]{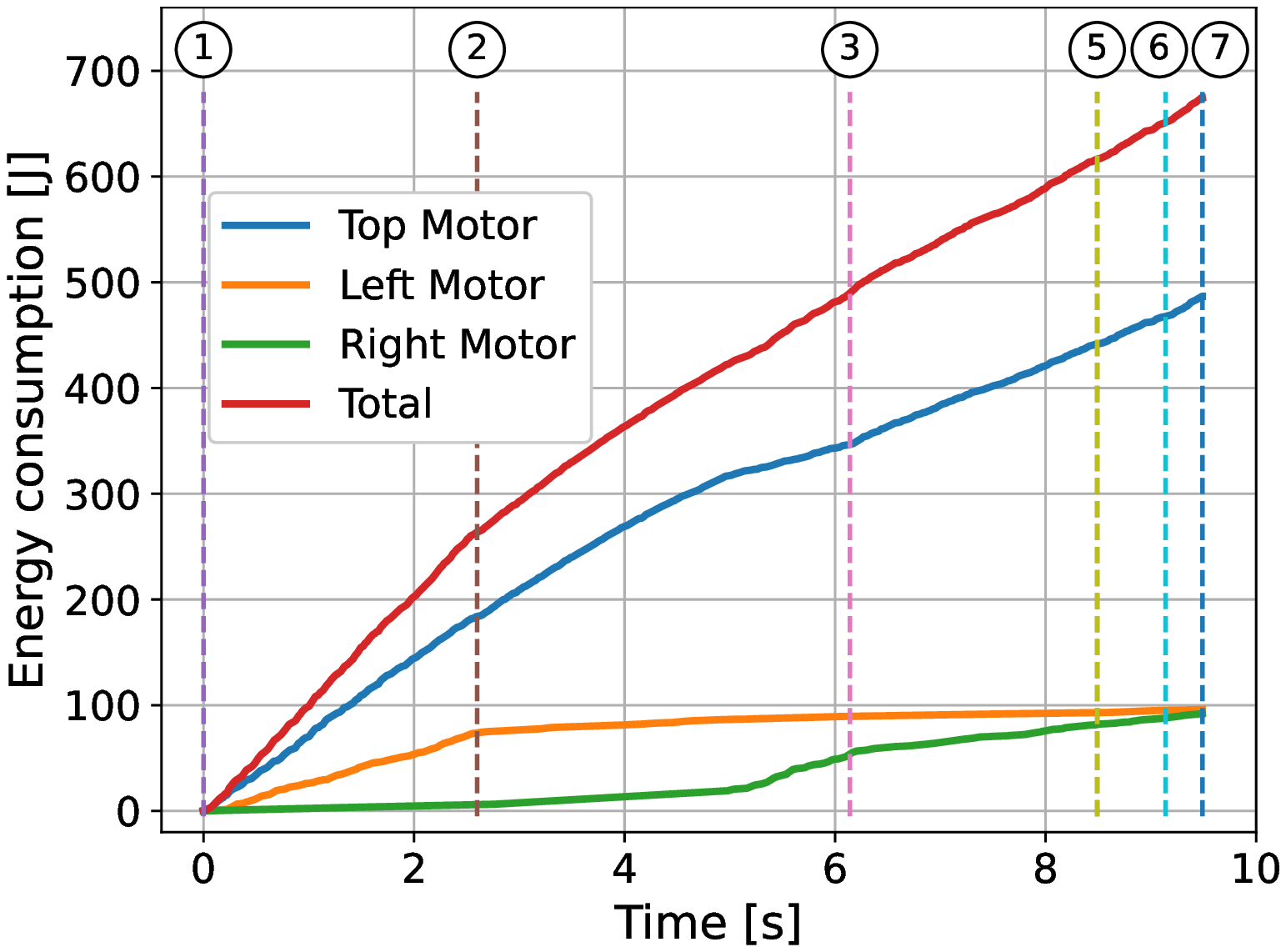}
        \caption{PTP controller}
        \label{fig:energy_PTP}
    \end{subfigure}
    \caption{Energy consumption corresponding to the trajectories 
    of Fig. \ref{fig:trajectories}, for each motor, as well as for 
    the whole system.
    }
    \label{fig:energy_consumption}
\end{figure*}

\noindent
The consumed energy is directly related to the total 
execution time. The lift and drop state are executed at the same 
speed, since these are similar and interchangeable parts in 
both trajectories. The other parts of the PTP-trajectory are  
scaled such that the same total execution time is achieved, 
and the maximum motor velocity is not exceeded. 
The energy consumption of each motor is given by eq. 
\eqref{eq:el_energy}, and is depicted in Fig. 
\ref{fig:energy_consumption}. Table \ref{table:energy} gives the 
consumed energy of each motor during each subtask, and the 
cumulative energy consumed by all motors.

\begin{table}[h]
\vspace{-4mm}
\caption{Energy consumption during each subtask, and the 
cumulative total.}
\label{table:energy} 
\centering
\setlength\tabcolsep{4.55pt}
\begin{tabular}{|l|l|c|c|c|c|c|c|}
    \hline
    \multicolumn{2}{|c|}{\multirow{2}{*}{}} &
    \multicolumn{6}{|c|}{Subtask energy consumption [J]} \\
    \cline{3-8}
    \multicolumn{2}{|c|}{} &
    \circledtable{1} \!$\rightarrow$\!\!\!
    \circledtable{2} 
    &
    \circledtable{2} \!$\rightarrow$\!\!\!
    \circledtable{3}  
    &
    \circledtable{3} \!$\rightarrow$\!\!\!
    \circledtable{4} 
    &
    \circledtable{4} \!$\rightarrow$\!\!\!
    \circledtable{5} 
    &
    \circledtable{5} \!$\rightarrow$\!\!\!
    \circledtable{6} 
    &
    \circledtable{6} \!$\rightarrow$\!\!\!
    \circledtable{7} \\
    \hline
    \multirow{2}{*}{Top} & 
    PTP & \raisebox{-.6pt}{183.87} & \raisebox{-.6pt}{162.81} & 
        \multicolumn{2}{|c|}{\raisebox{-.6pt}{94.76}} & 
        \raisebox{-.6pt}{26.46} & \raisebox{-.6pt}{18.80} \\  
    \cdashline{3-8}
    & Swing & \raisebox{-1.1pt}{199.06} & \raisebox{-1.1pt}{7.16} & 
        \raisebox{-1.1pt}{76.89} & \raisebox{-1.1pt}{27.97} & 
        \raisebox{-1.1pt}{9.10} & \raisebox{-1.1pt}{11.15} \\
    \hline
    \multirow{2}{*}{Left} & 
    PTP & \raisebox{-.6pt}{62.85} & \raisebox{-.6pt}{18.08} & 
        \multicolumn{2}{|c|}{\raisebox{-.6pt}{21.59}} & 
        \raisebox{-.6pt}{24.05} & \raisebox{-.6pt}{24.45} \\  
    \cdashline{3-8}
    & Swing & \raisebox{-1.1pt}{71.22} & \raisebox{-1.1pt}{0.17} & 
        \raisebox{-1.1pt}{0} & \raisebox{-1.1pt}{0} & 
        \raisebox{-1.1pt}{0} & \raisebox{-1.1pt}{0} \\
    \hline
    \multirow{2}{*}{Right} & 
    PTP & \raisebox{-.6pt}{5.90} & \raisebox{-.6pt}{47.77} & 
        \multicolumn{2}{|c|}{\raisebox{-.6pt}{27.95}} & 
        \raisebox{-.6pt}{6.40} & \raisebox{-.6pt}{3.99} \\  
    \cdashline{3-8}
    & Swing & \raisebox{-1.1pt}{6.27} & \raisebox{-1.1pt}{14.98} & 
        \raisebox{-1.1pt}{0.01} & \raisebox{-1.1pt}{3.56} & 
        \raisebox{-1.1pt}{0.15} & \raisebox{-1.1pt}{0.71} \\
    \hline
    \hline
    Total  & 
    PTP & \raisebox{-.6pt}{263.35} & \raisebox{-.6pt}{489.81} & 
        \multicolumn{2}{|c|}{\raisebox{-.6pt}{616.02}} & 
        \raisebox{-.6pt}{651.34} & \raisebox{-.6pt}{674.54} \\  
    \cdashline{3-8}
    (Cumulative)
    & Swing & \raisebox{-1.1pt}{276.54} & \raisebox{-1.1pt}{298.85} &
        \raisebox{-1.1pt}{375.74} & \raisebox{-1.1pt}{407.27} & 
        \raisebox{-1.1pt}{416.52} & \raisebox{-1.1pt}{428.38} \\
    \hline 
\end{tabular}
\vspace{-4mm}
\end{table}
\vspace{-4mm}

\noindent
During \circled{1} \!$\rightarrow$ \circled{2} energy is injected to 
overcome gravity. 
During \circled{2} \!$\rightarrow$ \circled{3} the biggest 
difference in energy consumption occurs. This is 
mainly because the top motor, which consumes the most power, is braked 
in the proposed method. In the proposed method no force is
generated in the left motor from this point onward. Either gravity 
produces tension in the corresponding cable for the rest of the motion, 
or the motion occurs in the half plane where this motor is disabled.

In this particular experimental setup, the weight of the payload cannot overcome the friction of 
the top motor, due to its high gear ratio. Thus, it still needs to be 
powered in order to drop the payload during \circled{3} 
$\!\rightarrow$ \circled{4} of the proposed method. 
However, by using the brakes of the right motor to control the drop, 
instead of actuating the motor, no power is consumed by that joint 
compared to the PTP method.

\begin{figure}[!h]
\vspace{-4mm}
\centering
\includegraphics[width = 7.2cm ]{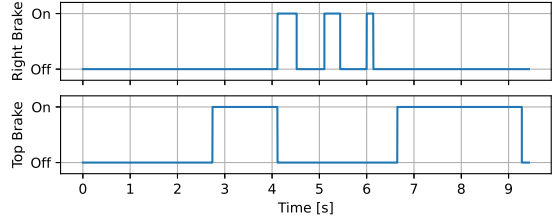}
\caption{Time during which each brake is active in the proposed 
        method. The left brake is omitted from this figure since 
        it is always off.
        } 
\label{fig:brake_time}
\vspace{-4mm}
\end{figure}

For the PTP controller, the brakes 
are continuously energised and consume a constant power. In the 
proposed control method, the brakes are only energised when it is 
necessary (Fig. \ref{fig:brake_time}). The consumed energy is calculated by 
multiplication of the on-time of the brake, with its 
power consumption (Table \ref{table:brake_energy}).

\vspace{-4mm}
\vspace{-4mm}

\begin{table}[!h]
\vspace{-4mm}
\centering
\setlength\tabcolsep{6pt}
\caption{Brake energy consumed for the 
    PTP- and the proposed controller.}
\label{table:brake_energy}
\begin{tabular}{|l|c|c|c||c|}
    \hline
    \multirow{2}{*}{} & \multicolumn{4}{|c|}{Brake energy consumption [J]}  \\ 
    \cline{2-5} 
    & Left & Right & Top & Total \\
    \hline
    PTP & 104.37 & 104.37 & 170.79 & 379.54 \\
    \hline
    Swing & 103.73 & 94.03 & 97.60 & 295.36 \\
    \hline
\end{tabular}
\vspace{-4mm}
\end{table}

\noindent
As indicated in Tables \ref{table:energy} and \ref{table:brake_energy}, 
the total energy consumed by the robot with the proposed approach is 
$723.74J$ compared to $1054.08 J$ of the PTP approach, which is a gain 
of more than $30\%$.
\section{Discussion and Conclusion}            \label{sec:conclusion}
In this paper a novel energy-efficient control method was introduced. 
This method is built upon simple principles, and only uses the 
kinematics of the system. It does not rely on complex models and 
does not need extensive quantitative identification. 
The dynamic effects in the motors and cables could be neglected.  
However, more effort is required to
specify the task in the control execution. While this might 
be beneficial for repetitive tasks with high cycles, for 
task with low cycles, the added 
effort might not outweigh the energy gains.  

Based on experimental results of a transportation task, 
the total energy consumption was $31\%$ 
lower with the proposed method compared to a conventional PTP controller,
on exactly the same hardware and software setup. 
The current implementation is not yet optimised, so it is expected 
that the energy can still be lowered if the execution is optimised.
This would however require knowledge of the dynamics of the system, 
and thus a more in-depth identification.
The principles of this method can be used for any task that have 
similar tolerances on path following and timing as the (subtasks) 
of the described use case.

The validation set-up was built to a large extend with low-cost 
hardware that was readily available. 
The jittery behaviour of the side motors (especially at low speeds) 
currently limit the achievable 
tracking behaviour.
But even with this non-ideal hardware, the task could be reliably 
executed.

In future work additional actuation 
will be added such that the top motor can be moved horizontally,
and the swing can occur wherever in the workspace, giving more 
flexibility in the execution.

\end{document}